\pdfoutput=1

\documentclass[11pt]{article}

\usepackage[]{acl}

\usepackage{times}
\usepackage{float}
\usepackage{graphicx}
\usepackage{latexsym}
\usepackage{natbib}
\usepackage{amsmath}
\usepackage{enumerate}
\usepackage{amssymb}
\usepackage{multirow}
\usepackage{multicol}
\usepackage{enumitem}
\usepackage{booktabs}
\usepackage{enumitem}
\usepackage{balance}
\usepackage{natbib}
\usepackage{url}
\usepackage[T1]{fontenc}
\usepackage[utf8]{inputenc}
\usepackage{microtype}
\newcommand{\ie}{\textit{i}.\textit{e}.}

\title{BIC: Twitter Bot Detection with Text-Graph \\ Interaction and Semantic Consistency}

\author{
    Zhenyu Lei$^1$\thanks{These authors contributed equally to this work.} \:
    Herun Wan$^1$\footnotemark[1]\:  Wenqian Zhang$^1$\:
    Shangbin Feng$^2$\\ \bf
    Zilong Chen$^3$ \:
    Jundong Li$^4$ \:
    Qinghua Zheng$^1$\:
    Minnan Luo$^1$\\
    Xi'an Jiaotong University$^1$, University of Washington$^2$ \\
    Tsinghua University$^3$, University of Virginia$^4$\\
    \texttt{\{Fischer, wanherun\}@stu.xjtu.edu.cn} \\
}

\begin{document}
\maketitle
\begin{abstract}
Twitter bots are automatic programs operated by malicious actors to manipulate public opinion and spread misinformation. Research efforts have been made to automatically identify bots based on texts and networks on social media. Existing methods only leverage texts or networks alone, and while few works explored the shallow combination of the two modalities, we hypothesize that the interaction and information exchange between texts and graphs could be crucial for holistically evaluating bot activities on social media. In addition, according to a recent survey \citep{cresci2020decade}, Twitter bots are constantly evolving while advanced bots steal genuine users' tweets and dilute their malicious content to evade detection. This results in greater inconsistency across the timeline of novel Twitter bots, which warrants more attention. In light of these challenges, we propose \textbf{BIC}, a Twitter \textbf{B}ot detection framework with text-graph \textbf{I}nteraction and semantic \textbf{C}onsistency. Specifically, in addition to separately modeling the two modalities on social media, BIC employs a text-graph interaction module to enable information exchange across modalities in the learning process. In addition, given the stealing behavior of novel Twitter bots, BIC proposes to model semantic consistency in tweets based on attention weights while using it to augment the decision process. Extensive experiments demonstrate that BIC consistently outperforms state-of-the-art baselines on two widely adopted datasets. Further analyses reveal that text-graph interactions and modeling semantic consistency are essential improvements and help combat bot evolution.
\end{abstract}

\section{Introduction}

\begin{figure}[t]
    \centering
    \includegraphics[width=1\linewidth]{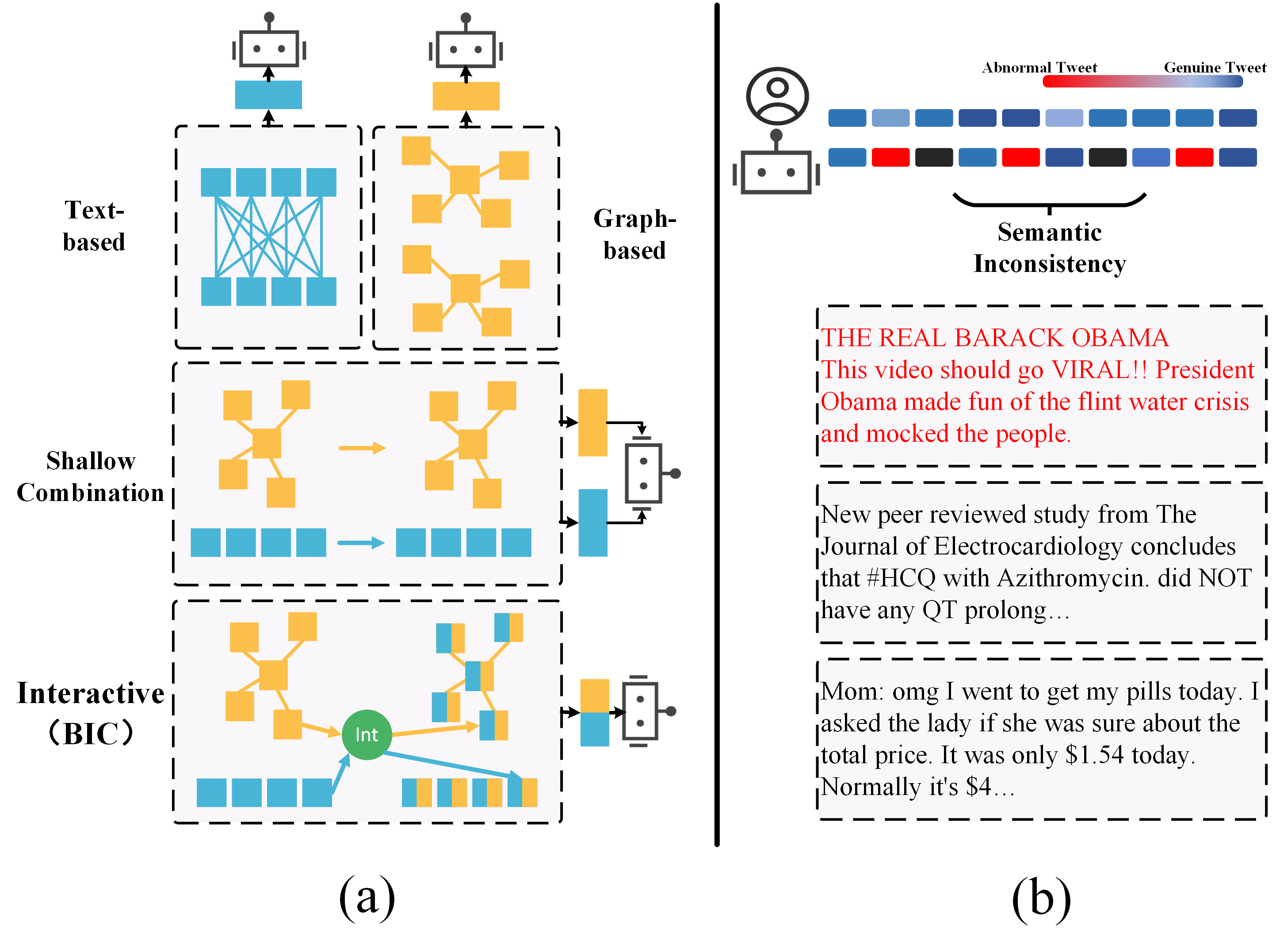}
    \caption{(a) Different types of combining modalities. Previous methods adopt text modality and graph modality alone, or just shallow combine them. There is a need for an interactive method that interacts and exchanges information across the modalities. (b) Genuine users and Twitter bots have different patterns of semantic consistency. Tweets in red are abnormal and these example tweets show semantic inconsistency.}
    \label{fig:teaser}
\end{figure}

Twitter bots are controlled by automated programs and manipulated to pursue malicious goals such as advocating for extremism and producing spam~\citep{dickerson2014using, berger2015isis}. Bots are also involved in spreading misinformation during the pandemic~\citep{shi2020social}. Since Twitter bots pose threat to online society, many efforts have been devoted to detecting bots.

The majority of the existing approaches are text-based and graph-based. The text-based methods analyze the content to detect Twitter bots by natural language processing techniques.~\citet{kudugunta2018deep} adopted recurrent neural networks to extract textual information.~\citet{guo2021socialbgsrd} utilized the pre-trained language model BERT to help detect bots. The graph-based methods model the Twittersphere as graphs and adopt geometric neural networks or concepts of network dynamics to identify bots.~\cite{feng2022heterogeneity} constructed a heterogeneous graph and leveraged the different relation information.~\citet{GraphHist} exploited the ego-graph of Twitter users and proposed a histogram and customized backward operator. 

However, existing methods are faced with two challenges. On the one hand, these methods only adopt texts or graphs alone, and only a few works shallowly combine the two modalities as Figure \ref{fig:teaser}(a) shows. The text-based model can not get the graph modality information while the graph-based model can not get the text modality information. We hypothesize that it is wise to interact and exchange information between texts and graphs to evaluate bot activities. On the other hand, \citet{cresci2020decade} pointed out that Twitter bots are constantly evolving. Advanced bots steal genuine users' tweets and dilute their malicious content to evade detection, which results in greater inconsistency across the timeline of advanced bots as Figure \ref{fig:teaser}(b) illustrates. Previous methods can not capture this characteristic. Namely, there is an urgent need for a method that can identify advanced bots.

In inspire of these challenges, we propose a framework BIC (Twitter \textbf{B}ot Detection with Text-Graph \textbf{I}nteraction and Semantic \textbf{C}onsistency). BIC separately models the two modalities, text and graph, in social media. A text module is adopted to encode the textual information and a graph module to encode graph information. BIC employs a text-graph interaction module to enable different modality information exchange across modalities in the learning process. To capture the inconsistency of advanced bots,  BIC leverages a semantic consistency module, which employs the attention weights and a sample pooling function. Our main contributions are summarized as follows:

\begin{figure*}
    \centering
    \includegraphics[width=0.9\linewidth]{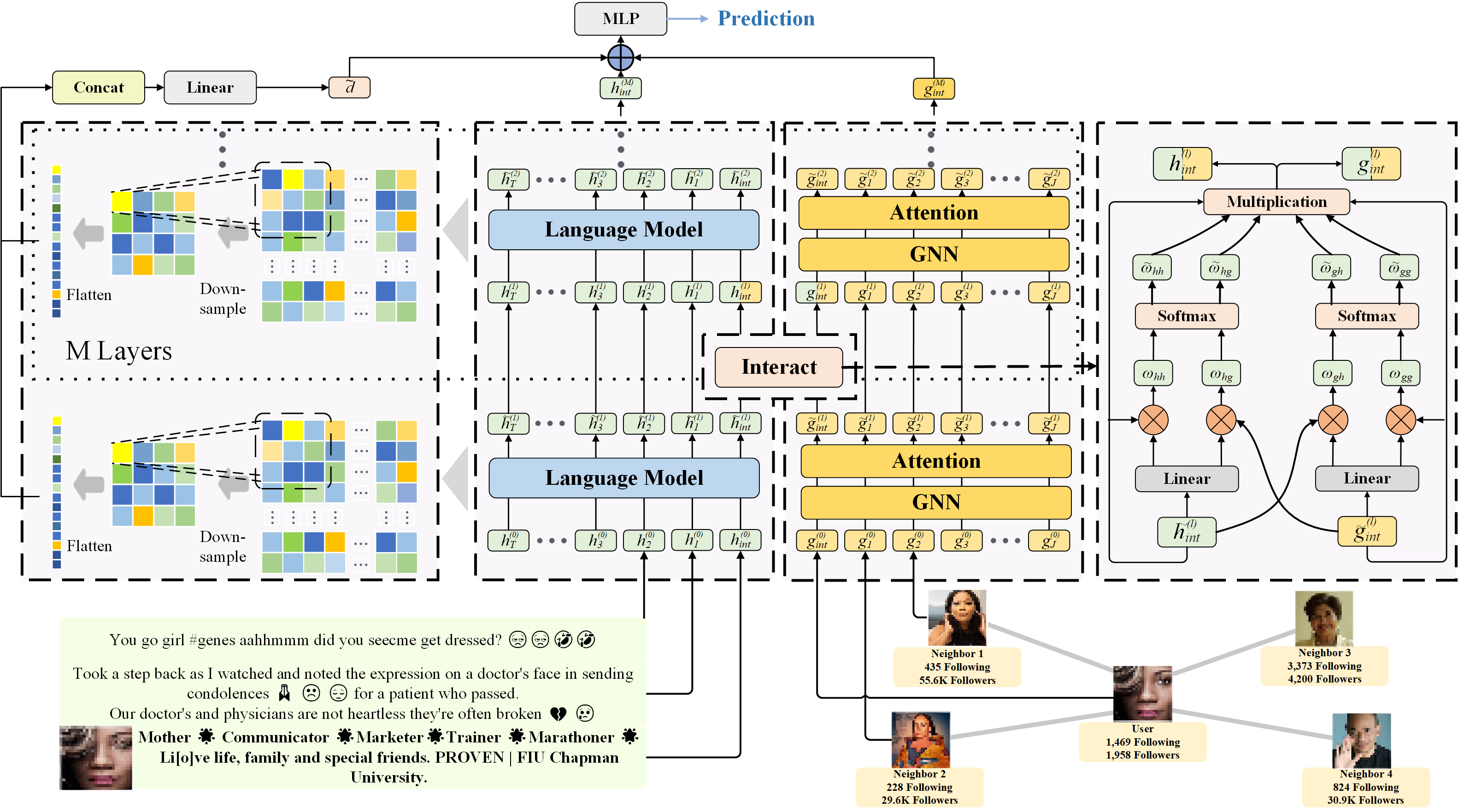}
    \caption{Overview of our proposed framework BIC.}
    \label{fig:overview}
\end{figure*}

\begin{itemize}[leftmargin=*]
    \item We propose to interact and exchange information across text and graph modalities to help detect bots. We find that capturing novel bots' inconsistency can increase detection performance.  

    \item  We propose a novel Twitter bot detection model, BIC. It is an end-to-end model and contains a text-graph interaction module to exchange modality information and a semantic consistency module to capture the inconsistency of advanced bots.

    \item We conduct extensive experiments to evaluate BIC and state-of-the-art models on two widely used datasets. Results illustrate that BIC outperforms all baseline methods. Further analyses reveal the effectiveness of the text-graph interaction module and semantic consistency module.

\end{itemize}

\section{Problem Definition}
We first define the task of Twitter bot detection with the text and graph modality. For a Twitter user $u_i\in U$, the text modality contains the description $B_i$ and the tweets $S_i=\{S_{i,j}\}_{j=1}^{T_i}$, where $T_i$ denotes the tweet count. The graph modality contains the representation $f_i$ of $u_i$ and the heterogeneous graph $\mathcal{G} = 
\mathcal{G}(U, E, \varphi, R^e)$, where $U$ denotes the user set, $E$ denotes the edge set, $\varphi: E\xrightarrow{}R^e$ denotes the relation mapping function and $R^e$ is the relation type set. The neighbors of $u_i$ can be derived from $\mathcal{G}$ as $N_i=\{n_{i,j}\}_{j=1}^{J_i}$ where $J_i$ is the neighbor count. The goal is to find a detection function $f: f(u_i)\xrightarrow{}\hat{y}\in\{0, 1\}$, such that $\hat{y}$ approximates ground truth $y$ to maximize prediction accuracy.

\section{Methodology}
Figure~\ref{fig:overview} displays an overview of our proposed framework named BIC. Specifically, BIC firstly leverages a text module to encode textual information and a graph module to encode graph information. BIC then adopts a text-graph interaction module to interact and exchange modality information in the learning process. To further interact the two modalities, BIC repeats this process for $M$ times. BIC extracts the semantic consistency from the attention weights from the text module with the help of the semantic consistency module. Finally, BIC leverages text modality, graph modality, and semantic consistency vectors to identify bots.

\subsection{Modality Interaction}
For simplicity, we omit the subscript of the user. BIC first encodes the text modality and graph modality information to obtain the initial representations. For text modality, BIC employs pre-trained RoBERTa~\citep{DBLP:journals/corr/abs-1907-11692} to encode description $B$ and tweets $\{S\}_{i=1}^T$ into $h^{(0)}_{int}$ and $\{h^{(0)}_i\}_{i=1}^T$. BIC considers $h^{(0)}_{int}$ as the text interaction modality because the description generally defines the user. For graph modality, BIC employs the same encoding methods as BotRGCN~\citep{feng2021botrgcn} to get the graph interaction representation $g_{int}^{(0)}$ and representations of its neighbors $\{g_{i}^{(0)}\}_{i=1}^{J}$.

After obtaining the initial representations, BIC employs $M$ times modality interaction to ensure text and graph information interact completely. We describe the $l$-th interact process as follows.

\paragraph{Text Module}BIC puts text representations into a language model to extract textual information, \ie,
\begin{equation}
\label{equ:text}
\resizebox{\linewidth}{!}{
    $\{\tilde{h}^{(l)}_{int}, \tilde{h}^{(l)}_1, \cdots, \tilde{h}^{(l)}_T\} = {\rm LM}(\{h^{(l-1)}_{int}, h^{(l-1)}_1, \cdots, h^{(l-1)}_T\}),$}
\end{equation}
where $\tilde{h}^{(l)}_{int}$ denotes interaction representation of text modality before interaction. BIC adopts transformer with multi-head attention~\citep{vaswani2017attention} as the language model ${\rm LM}$.

\paragraph{graph module}BIC firstly feeds graph representations into a graph neural network to aggregate information between users and its neighbors, \ie,
\begin{equation*}
\resizebox{\linewidth}{!}{$
    \{\hat{g}^{(l)}_{int}, \hat{g}^{(l)}_1, \cdots, \hat{g}^{(l)}_J\} = {\rm GNN}(\{g^{(l-1)}_{int}, g^{(l-1)}_1, \cdots, g^{(l-1)}_J\}).
$}
\end{equation*}
BIC adopts relational graph convolutional networks~\citep{schlichtkrull2018modeling} due to its ability to extract heterogeneous information. To measure which neighbor is important for bot detection, BIC employs multi-head attention for the user, \ie,
\begin{equation*}
\resizebox{\linewidth}{!}{$
    \{\tilde{g}^{(l)}_{int}, \tilde{g}^{(l)}_1, \cdots, \tilde{g}^{(l)}_J\} = {\rm att}(\{\hat{g}^{(l)}_{int}, \hat{g}^{(l)}_1, \cdots, \hat{g}^{(l)}_J\}),
$}
\end{equation*}
where $\tilde{g}^{(l)}_{int}$ denotes interaction representation of graph modality before interaction and ${\rm att}$ denotes multi-head attention.

\subsubsection{Text-Graph Interaction Module}
\label{sec:4.1.3}
BIC adopts a text-graph interaction module to interact and exchange information across text and graph modality in the learning process. Specifically, BIC employ a interaction function ${\rm inter}$ to interact the text modality representation $\tilde{h}^{(l)}_{int}$ and the graph modality representation $\tilde{g}^{(l)}_{int}$, \ie,
\begin{equation*}
    (g_{int}^{(l)}, h_{int}^{(l)}) = {\rm inter}(\tilde{g}_{int}^{(l)}, \tilde{h}_{int}^{(l)}).
\end{equation*}

For the detail about ${\rm inter}$ function, BIC calculates the similarity coefficient between modality representations, \ie,
\begin{equation}
    \begin{aligned}
        w_{hh} = \tilde{h}_{int}^{(l)} \otimes (\theta_1\cdot\tilde{h}_{int}^{(l)}),\\
        w_{hg} = \tilde{h}_{int}^{(l)} \otimes (\theta_2\cdot\tilde{g}_{int}^{(l)}),\\
        w_{gg} = \tilde{g}_{int}^{(l)} \otimes (\theta_2\cdot\tilde{g}_{int}^{(l)}), \\
        w_{gh} = \tilde{g}_{int}^{(l)} \otimes (\theta_1\cdot\tilde{h}_{int}^{(l)}),
    \end{aligned}
    \label{eq:s_weight}
\end{equation}
where $\theta_1$ and $\theta_2$ are learnable parameters that transform the modality representations into the interaction-sensitive space, and `$\otimes$' denotes the dot product. BIC then applies a softmax function to derive final similarity weights, \ie,
\begin{equation*}
\begin{aligned}
        \tilde{w}_{hh}, \tilde{w}_{hg} = {\rm softmax}(w_{hh}, w_{hg}),\\
        \tilde{w}_{gg}, \tilde{w}_{gh} = {\rm softmax}(w_{gg}, w_{gh}).
\end{aligned}
\end{equation*}

BIC finally makes the two representations interact through the derived similarity weights, \ie,
\begin{equation*}
\begin{aligned}
        h_{int}^{(l)} = \tilde{w}_{hh} \tilde{h}_{int}^{(l)} + \tilde{w}_{hg}  \tilde{g}_{int}^{(l)},\\
        g_{int}^{(l)} = \tilde{w}_{gg} \tilde{g}_{int}^{(l)} + \tilde{w}_{gh} \tilde{h}_{int}^{(l)}.\\
\end{aligned}
\end{equation*}
So far, BIC could interact and exchange information across the two modalities.

\subsection{Semantic Consistency Detection}
Since attention weights from the transformer could indicate the correlations and consistency between tweets, BIC adopts the attention weights to extract the semantic consistency information. BIC can obtain the attention weight matrix $\mathcal{M}_i \in\mathbb{R}^{(T+1) \times (T+1)}$ of text representation from equation~(\ref{equ:text}) in $i$-th interaction process. BIC then employs a down-sample function to reduce the matrix size and obtain what matters in the matrix, \ie, 
\begin{equation*}
    \tilde{\mathcal{M}}_i = {\rm sample}(\mathcal{M}_i), \  \tilde{\mathcal{M}}_i\in\mathbb{R}^{K\times K},
\end{equation*}
where $K$ is a hyperparameter indicating the matrix size. BIC adopts fixed size max-pooling as ${\rm sample}$ function in the experiments. BIC then flat the matrix and applies a linear transform to obtain the semantic consistency representation, \ie,
\begin{equation*}
    d_i = \theta_{sc}\cdot {\rm Flatten}(\tilde{\mathcal{M}}_i),
\end{equation*}
where $\theta_{sc}$ is a shared learnable parameter of each interaction process. Finally, BIC applies an aggregating function to combine the representations of each interaction process, \ie,
\begin{equation*}
    d = \sigma(W_D\cdot {\rm aggr}(\{d_i\}_{i=1}^M) + b_D),
\end{equation*}
where $W_D$ and $b_D$ are learnable parameters, $\sigma$ denotes activate function, and ${\rm aggr}$ denotes the aggregating function, such as ${\rm concatenate}$ or ${\rm mean}$.

\subsection{Training and Inference}
BIC concatenates text modality $h_{int}^{(M)}$, graph modality $g_{int}^{(M)}$, and semantic consistency $d$ representation to obtain the representation of a user, \ie,
\begin{equation}
    \label{eq:rep}
    z = W_D\cdot (h_{int}^{(M)}\|g_{int}^{(M)}\|d) + b_D.
\end{equation}
BIC finally employs a softmax layer to get the predicted probability $\hat{y}$. We adopt cross entropy loss to optimize BIC, \ie,
\begin{equation*}
\resizebox{\linewidth}{!}{$\begin{aligned}
    l = -\sum_{i\in U}[y_i\log(\hat{y_i}) + (1 - y_i)\log(1 - \hat{y_i})] + \lambda\sum_{\omega\in\theta}\omega^{2},
\end{aligned}$}
\end{equation*}
where $U$ denotes all users in the training set, $\theta$ denotes all training parameters, $y_i$ denotes the ground-truth label and $\lambda$ is a regular coefficient.

\begin{table*}[h]
    \centering
    \caption{Bot detection performance on Cresci-15 and TwiBot-20 benchmarks. For each baseline except for Botometer which has fixed results, we run 5 times on the same splits with different random seeds. \textbf{Text}, \textbf{Graph}, \textbf{Modality-Int} respectively denote whether baseline leverages text modality, graph modality and modality interaction. \textbf{Bold} and \underline{underline} indicate the highest and second highest performance. `BIC w/o Graph' and `BIC w/o Text' indicate BIC without the Graph Module and without the Text Module. BIC achieves the best performance.}
        \resizebox{\textwidth}{!}{
        \begin{tabular}[c]{lccccccc}
            \toprule[1.5pt]
            \multirow{2}{*}{\textbf{Method}} & \multicolumn{3}{c}{\textbf{Modalities}} & \multicolumn{2}{c}{\textbf{Cresci-15}}& \multicolumn{2}{c}{\textbf{TwiBot-20}} \\
            \cmidrule[0.75pt](lr){2-4}
            \cmidrule[0.75pt](lr){5-6}
            \cmidrule[0.75pt](lr){7-8}
            & \textbf{Text} & \textbf{Graph} & \textbf{Modality-Int}& \textbf{Accuracy} & \textbf{F1-score}& \textbf{Accuracy} & \textbf{F1-score} \\
            \midrule[0.75pt] 
    		Yang \textit{et al.} & & && $77.08~(\pm 0.21)$& $77.91~(\pm 0.11)$& $81.64~(\pm 0.46)$& $84.89~(\pm 0.42)$\\
    		Botometer & &&& $57.92$& $66.90$ & $53.09$& $55.13$\\
    		Kudugunta \textit{et al.} &$\checkmark$ && & $75.33~(\pm 0.13)$& $75.74~(\pm 0.16)$& $59.59~(\pm 0.65)$& $47.26~(\pm 1.35)$\\
    		Wei \textit{et al.} &$\checkmark$ & & & $96.18~(\pm 1.54)$&$82.65~(\pm 2.47)$&$70.23~(\pm 0.10)$ &$53.61~(\pm 0.10)$ \\
    		BotRGCN &  &$\checkmark$ & & $96.52~(\pm 0.71)$ & $97.30~(\pm 0.53)$& $83.27~(\pm 0.57)$& $85.26~(\pm 0.38)$\\
    		Alhossini \textit{et al.}&  &$\checkmark$ & & $89.57~(\pm 0.60)$& $92.17~(\pm 0.36)$& $59.92~(\pm 0.68)$& $72.09~(\pm 0.54)$\\
    		RGT &  & $\checkmark$& & $97.15~(\pm 0.32)$ & $97.78~(\pm 0.24)$& $\underline{86.57}~(\pm 0.41)$ & $\underline{88.01}~(\pm 0.41)$\\
            SATAR& \checkmark&\checkmark& & $93.42~(\pm 0.48)$& $95.05~(\pm 0.34)$& $84.02~(\pm 0.85)$& $86.07~(\pm 0.70)$\\
    		\midrule[0.75pt]
    		BIC w/o Graph & $\checkmark$ & & &$\underline{97.16}~(\pm 0.58)$ & $\underline{97.80}~(\pm 0.46)$ & $85.44~(\pm 0.32)$ & $86.97~(\pm 0.41)$ \\
    		BIC w/o Text &  & $\checkmark$ &  &$96.86~(\pm 0.52)$ &$97.57~(\pm 0.39)$&$85.78~(\pm 0.48)$ &$87.25~(\pm 0.57)$   \\
    		\textbf{BIC} & $\checkmark$ & $\checkmark$ & $\checkmark$ & $\textbf{98.35}~(\pm 0.24)$ &$\textbf{98.71}~(\pm 0.18)$& $\textbf{87.61}~(\pm 0.21)$ &$\textbf{89.13}~(\pm 0.15)$ \\
            \bottomrule[1.5pt]
        \end{tabular}
        }
    \label{tab:main}
\end{table*}

\section{Experiment}
\subsection{Experiment Settings}
More detailed information about the experiment settings and the implementation details of BIC can be found in the appendix. we submit our code and the best parameters as supplementary materials.
\paragraph{Dataset}
To evaluate BIC and baselines, we make use of two widely used datasets, Cresci-15~\citep{cresci2015fame} and TwiBot-20~\citep{feng2021twibot}. These two datasets provide user follow relationships to support graph-based models. TwiBot-20 includes $229,580$ Twitter users, $33,488,192$ tweets, $33,716,171$ edges while Cresci-15 includes $5,301$ Twitter users, $2,827,757$ tweets, $14,220$ edges.

\paragraph{Baselines}
We compare BIC with \textbf{Botometer} \citep{davis2016botornot}, \textbf{Kudugunta \textit{et al.}} \citep{kudugunta2018deep}, \textbf{Wei \textit{et al.}}~\citep{wei2019twitter}, \textbf{Alhosseini \textit{et al.}}~\citep{ali2019detect}, \textbf{BotRGCN}~\citep{feng2021botrgcn}, \textbf{Yang \textit{et al.}}~\citep{yang2020scalable}, \textbf{SATAR}~\citep{feng2021satar}, and \textbf{RGT}~\citep{feng2022heterogeneity}. 

\subsection{Main Results}

We first evaluate whether these methods leverage text modality, graph modality, and interact modalities. We then benchmark these baselines on Crescie-15 and TwiBot-20, and present results in Table~\ref{tab:main}. It is demonstrated that:
\begin{itemize}[leftmargin=*]
    \item BIC consistently outperforms all baselines including the state-of-art methods RGT~\citep{feng2022heterogeneity} with at least $1\%$ improvement of performance on two datasets.
    \item The methods leveraged graph modality such as RGT~\citep{feng2022heterogeneity} generally outperform other methods that only adopt text modality or other features. SATAR~\citep{feng2021satar} achieves competitive performance with the text modality and the graph modality. BIC further makes these two modalities interact to achieve the best performance.
    \item We conduct the significance test using the unpaired t-test. The improvement between BIC and the second-best baseline RGT is statistically significant with p-value < 0.005 on Creaci-15 and p-value < 0.0005 on TwiBot-20. 
\end{itemize}

In the following, we first study the role of the two modalities and the interaction module in BIC. We then examine the effectiveness of the semantic consistency module in identifying advanced bots. Next, we evaluate the ability of BIC to detect advanced bots. We finally evaluate a specific bot in the datasets to explore how BIC makes the choice.

\subsection{Text-Graph Interaction Study}
\paragraph{Modality Effectiveness Study} We remove the text modality representation $h_{int}^{(M)}$ and the graph modality representation $g_{int}^{(M)}$ in equation~(\ref{eq:rep}), to evaluate the role of each modality. The results are illustrated in Table~\ref{tab:main}. We can conclude that: (i) Removing any modality will cause a drop in performance, which illustrates that leveraging and making the two modalities interact can help identify bots. (ii) BIC without graph modality can achieve the second-best performance on Cresci-15. Other ablation settings can achieve competitive performance. It is shown that BIC can derive useful information from one modality and the semantic consistency representation can help identify bots.

\begin{figure}[t]
    \centering
    \includegraphics[width=0.9\linewidth]{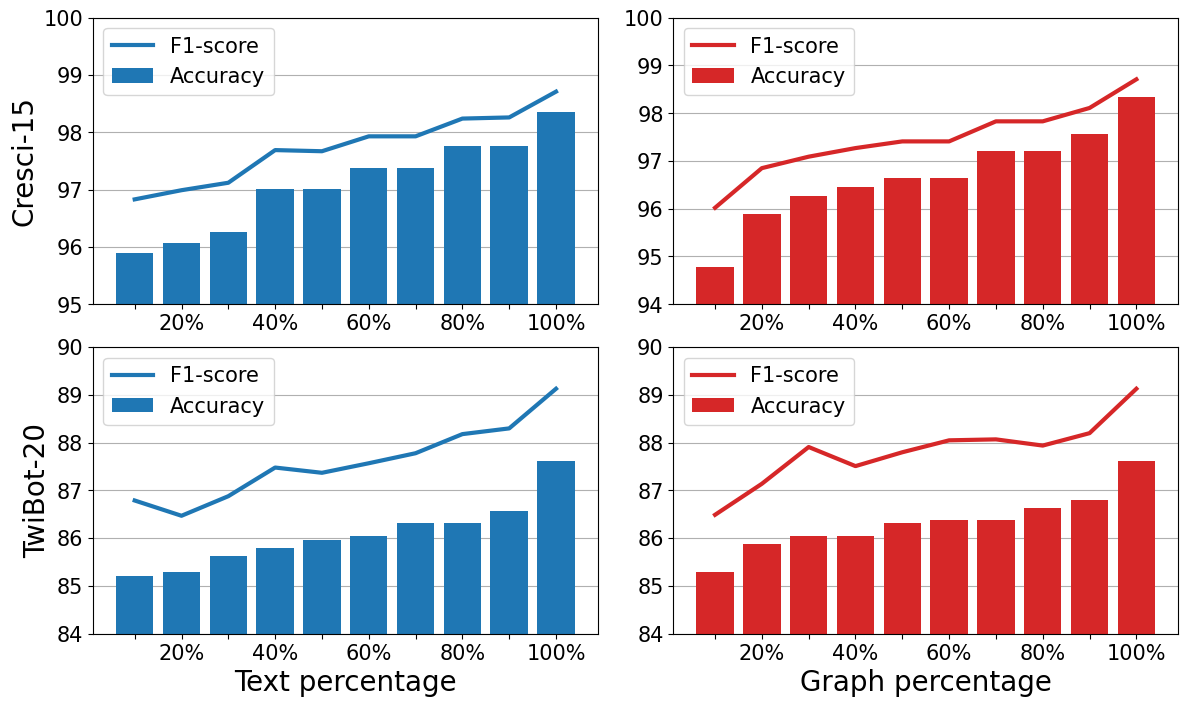}
    \caption{The performance of BIC trained with data that part of one modality is gradually removed. The results illustrate that every modality information benefits the performance and BIC could keep the performance with less information of one modality.}
    \label{fig:modality}
\end{figure}

BIC adopts text and graph modalities and leverages the text-graph interaction module to make information across the two modalities. To further examine the ability of BIC to extract modality information, we gradually remove part of one modality information and conduct experiments. The results in Fig~\ref{fig:modality} demonstrate that: (i) Every modality information benefits the performance of bot detection. It suggests that bot detection relies on the text modality and the graph modality information. (ii) BIC could keep the performance with less information of one modality. It illustrates that the interaction module is effective in exchanging information across the modalities.

\paragraph{Interaction Function Study} BIC employs an interaction function, which transforms representations into an interaction-sensitive space and learns the similarity weights, to exchange the modality information. Apart from our proposed similarity-based interaction, there are several other interaction functions. We replace this function with other functions such as mean or MLP, to evaluate the effectiveness of our proposed interaction function. We apply the following different interaction functions:
\begin{itemize}[leftmargin=*]
    \item\textbf{Hard} function computes the average of two interaction representations to interact.
    \item\textbf{Soft} function utilizes two learnable parameters as weights for two interaction representations to generate new representations.
    \item\textbf{MLP} function concatenates two interaction representations and feeds the intermediate into an MLP layer to interact.
    \item\textbf{Text} function feeds the interaction representation from text modality into Linear layers.
    \item\textbf{Graph} function feeds the interaction representation from graph modality into Linear layers.
\end{itemize}

The results in Table~\ref{tab:int} illustrate that:

\begin{itemize}[leftmargin=*]\setlength{\itemsep}{0pt}
    \item Almost all interaction strategies outperform methods with no interaction, which indicates the necessity of utilizing an interaction module to make two modalities interactive and exchange information. 
    \item Our similarity-based modality interaction function outperforms others all, which well confirmed its efficacy, indicating that it can truly make two modalities inform each other and learn the relative importance of modalities.
\end{itemize} 

\begin{table}[t]
    \centering
     \caption{Performance of model with different interaction functions. The results illustrate the effectiveness of the proposed similarity-based interaction.}
    \resizebox{\linewidth}{!}{
    \begin{tabular}{lcccc}
         \toprule[1.5pt]
         \multirow{2}{*}{\textbf{Function}} & \multicolumn{2}{c}{\textbf{Cresci-15}} & \multicolumn{2}{c}{\textbf{TwiBot-20}}\\
         \cmidrule[0.75pt](lr){2-3}
         \cmidrule[0.75pt](lr){4-5}
         & \textbf{Accuracy} & \textbf{F1-score} & \textbf{Accuracy} & \textbf{F1-score}\\
         \midrule[0.75pt]
         \textbf{Ours} &$\textbf{98.35}$& $\textbf{98.71}$& $\textbf{87.61}$& $\textbf{89.13}$\\
         w/o interaction &$95.89$&$96.85$& $85.97$& $87.42$\\
         Hard & $96.64$ & $97.41$& $86.64$& $88.15$ \\
         Soft & $97.01$ & $97.69$& $87.06$ & $88.27$ \\
         MLP &  $97.38$ & $97.97$&$86.98$& $88.44$ \\
         Text & $96.64$ & $97.41$& $85.63$& $87.14$ \\
         Graph &$96.45$ &$97.27$ & $86.30$& $87.65$ \\
         \bottomrule[1.5pt]
    \end{tabular}
    }
    \label{tab:int}
\end{table}

\begin{figure}[t]
    \centering
    \includegraphics[width=0.9\linewidth]{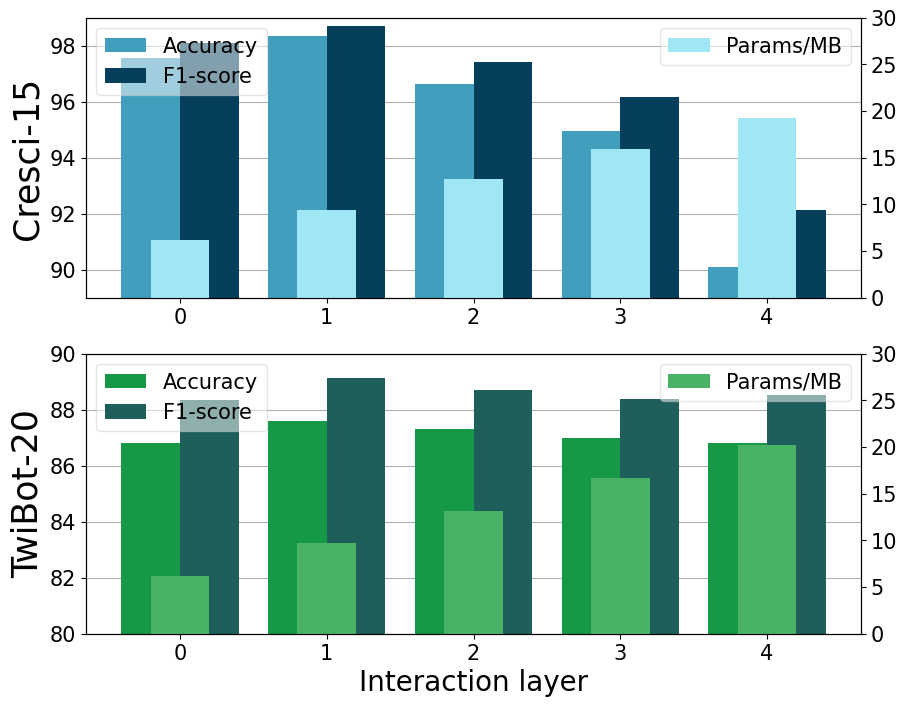}
    \caption{Performance of different numbers of model interaction layers and Params used for one training epoch. The results illustrate that model with 1 interaction layer has good performance with relatively lower Params.}
    \label{fig:layer}
\end{figure}

\paragraph{Interaction Number Study} To examine the role of the modality information interaction number $M$, we conduct experiments with different interaction numbers and evaluate the model memory cost~(Params). The results in Figure~\ref{fig:layer} demonstrate that BIC with 2 interactions performs the best over other settings. Besides, the two-layer interaction model has relatively less memory cost, which makes it the best selection. As the number of interaction number increases, the performance declines gradually, which may be caused by higher complexity increasing the training difficulty. Meanwhile, the one-layer interaction model may be deficient for learning so rich information, thus leading to unappealing performance. 

\subsection{Semantic Consistency Study}

\begin{figure}
    \centering
    \includegraphics[width=0.8\linewidth]{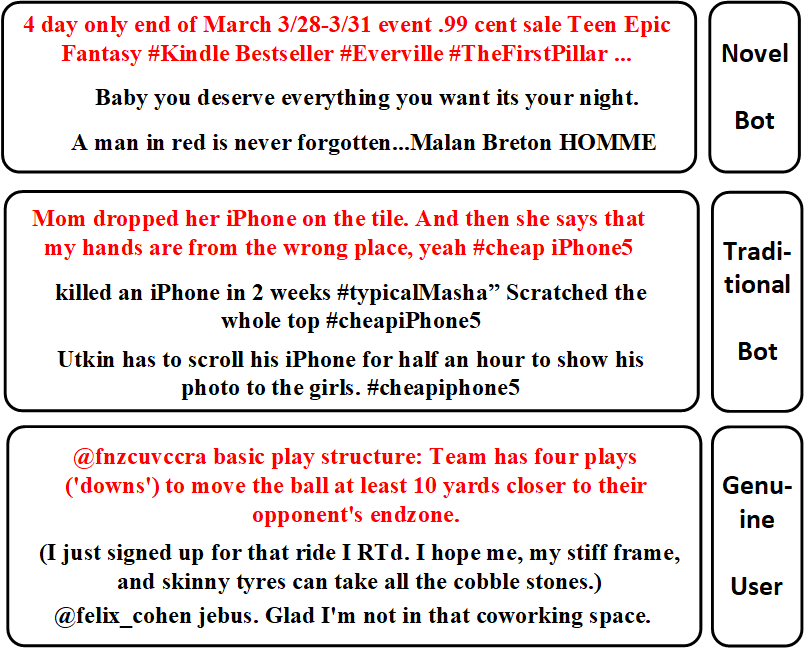}
    \caption{Representative tweets of a genuine user, a traditional bot, and an advanced bot. The tweet in red indicates it has a relatively higher attention weight than other tweets of the same user. More inconsistency has been shown between the advanced bot's tweets in red and tweets in black. }
    \label{fig:case_sc_}
\end{figure}

\begin{figure}
    \centering
    \includegraphics[width=1\linewidth]{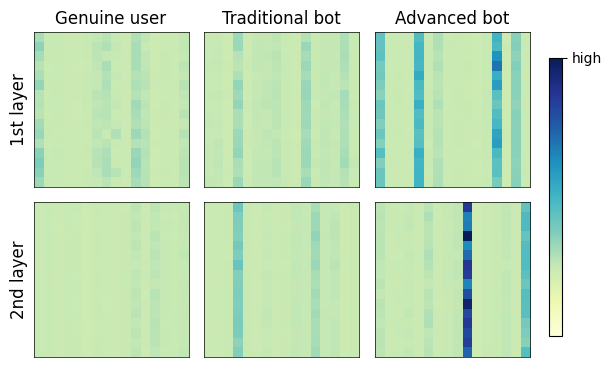}
    \caption{Semantic consistency matrices of a genuine user, a traditional bot, and an advanced bot. The result illustrates that the matrices of advanced bots show more inconsistency than traditional ones or humans.}
    \label{fig:case_sc}
\end{figure}

\paragraph{Discrimination Case Study}
We check the tweets of users in the used datasets to determine that humans and bots have different semantic consistency patterns and that advanced bots may steal genuine tweets. We choose a genuine user, a traditional bot, and an advanced bot. Their representative tweets are displayed in Figure~\ref{fig:case_sc_} and we can find that novel bots will have more inconsistency in tweets than genuine users and traditional bots which posts similar rubbish tweets. Next, we check their semantic consistency matrices $\tilde{\mathcal{M}}_i$ and they are shown in Figure~\ref{fig:case_sc}. We can find that the advanced bot has relatively higher inconsistency in its matrices.

\begin{figure}[t]
    \centering
    \includegraphics[width=1\linewidth]{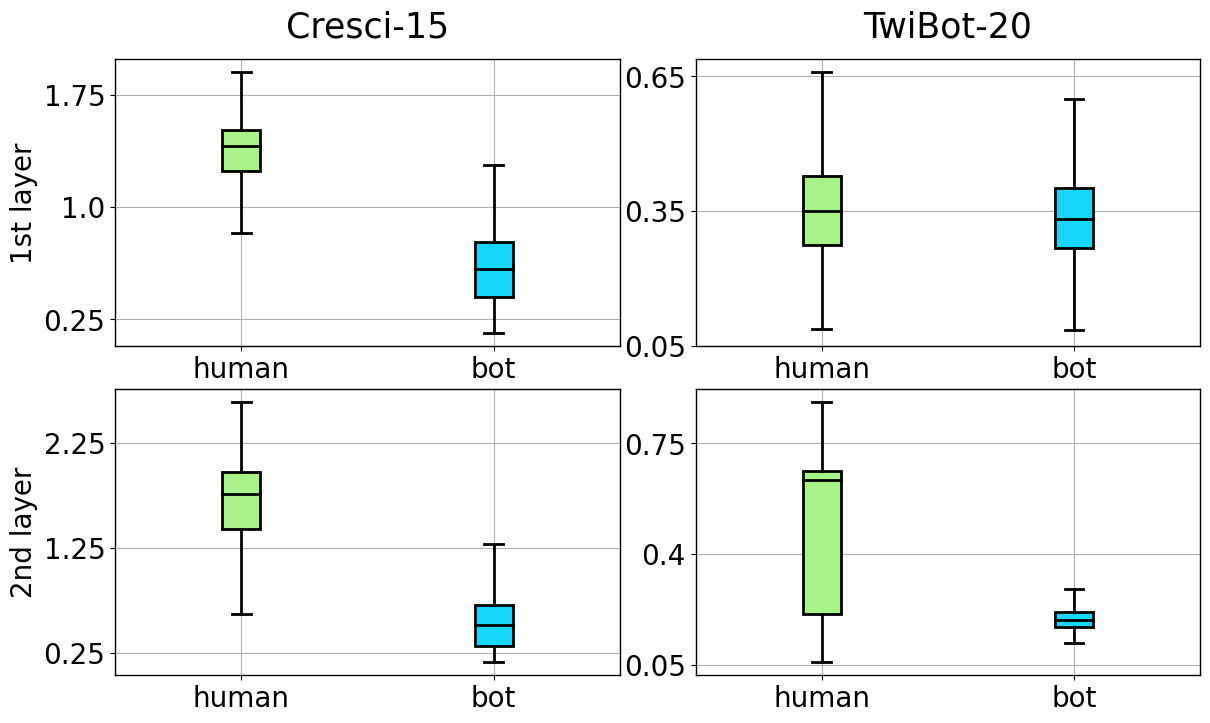}
    \caption{The box plot is drawn from the max characteristic values of the semantic consistency matrices. The results illustrate that the consistency matrices of humans and bots show different patterns.}
    \label{fig:box}
\end{figure}

\paragraph{Discrimination Ability Study}
BIC adopts the attention weight from the text module to generate the semantic consistency representation $d$. We try to find out the extent to which our semantic consistency module can distinguish bots from genuine users. We derive consistency matrices $\tilde{\mathcal{M}}_i$ and calculate the largest characteristic value. We draw box plots with these characteristic values to find the differences between bots and humans excavated by the module. The results manifested in Figure~\ref{fig:box} demonstrate that the consistency matrices of bots and humans exist in differences. 

\begin{figure}
    \centering
    \includegraphics[width=1\linewidth]{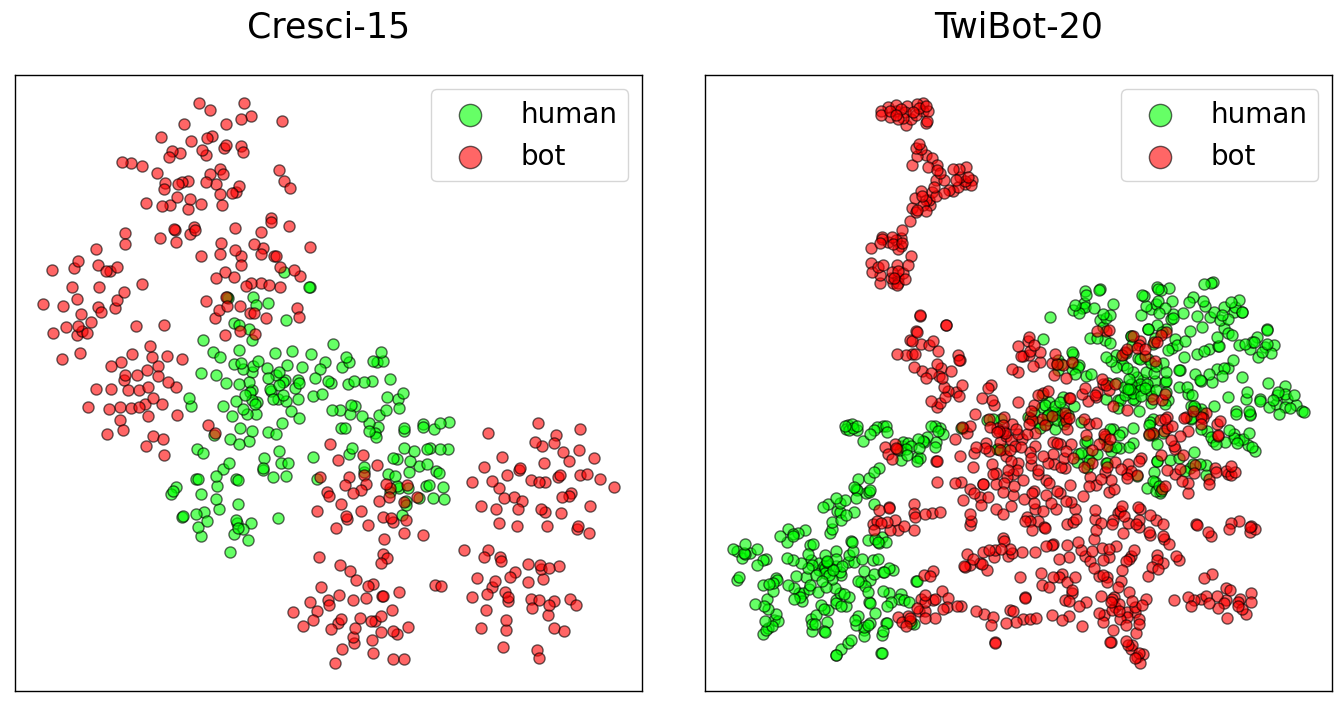}
    \caption{The t-sne plot of the semantic consistency representations. The results illustrate that the representation of humans and the representations of bots are obviously separated, which indicates the effectiveness of the semantic consistency module.}
    \label{fig:tsne}
\end{figure}

To evaluate that the semantic consistency representation $d$ can distinguish bots and humans. We conduct the k-means algorithm to cluster the representations and calculate the V-measure, which is a harmonic mean of homogeneity and completeness. BIC achieves 0.4312 of v-measure on Cresci-15 and 0.3336 on TwiBot-20. More intuitively, we adopt t-sne to visualize the representation and the results are shown in Figure~\ref{fig:tsne}, which shows moderate collocation for groups of bot and human. It is proven that the semantic consistency representation can identify bots alone.
\subsection{Advanced Bot Study}
We claim that BIC could identify the advanced bots. To evaluate whether BIC can capture the advanced bots after 2020 (the TwiBot-20 published time), we sample some users related to the pandemic from a new Twitter crawl~\citep{feng2022twibot} to construct a new dataset. This dataset contains user-follow relationships including 5,000 humans and 5,000 bots. We compare BIC with RGT, the second-best baseline, and Botometer, the widely-used bot detection tool. We randomly split this dataset into the train set and the test set by 8:2 and train the methods. Table~\ref{tab:covid} illustrates the results. We can conclude that BIC achieves the best performance, which proves that BIC can capture advanced bots with the help of the text-graph interaction module and the semantic consistency module.
\begin{table}
    \centering
    \caption{Bot detection performance on an up-to-date dataset. BIC outperforms the other two baselines, which illustrates BIC can better identify advanced bots.}
    \begin{tabular}{lcc}
        \toprule[1.5pt]
        \textbf{Method} &\textbf{Accuracy} &\textbf{F1-score} \\
        \midrule[0.75pt]
        Botometer & $55.35$& $53.99$\\
        RGT &$66.95$& $64.48$\\
        BIC &$\textbf{67.25}$&$\textbf{67.78}$\\
        \bottomrule[1.5pt]
    \end{tabular}
    
    \label{tab:covid}
\end{table}

\subsection{Case Study}
\begin{figure}[t]
    \centering
    \includegraphics[width=1\linewidth]{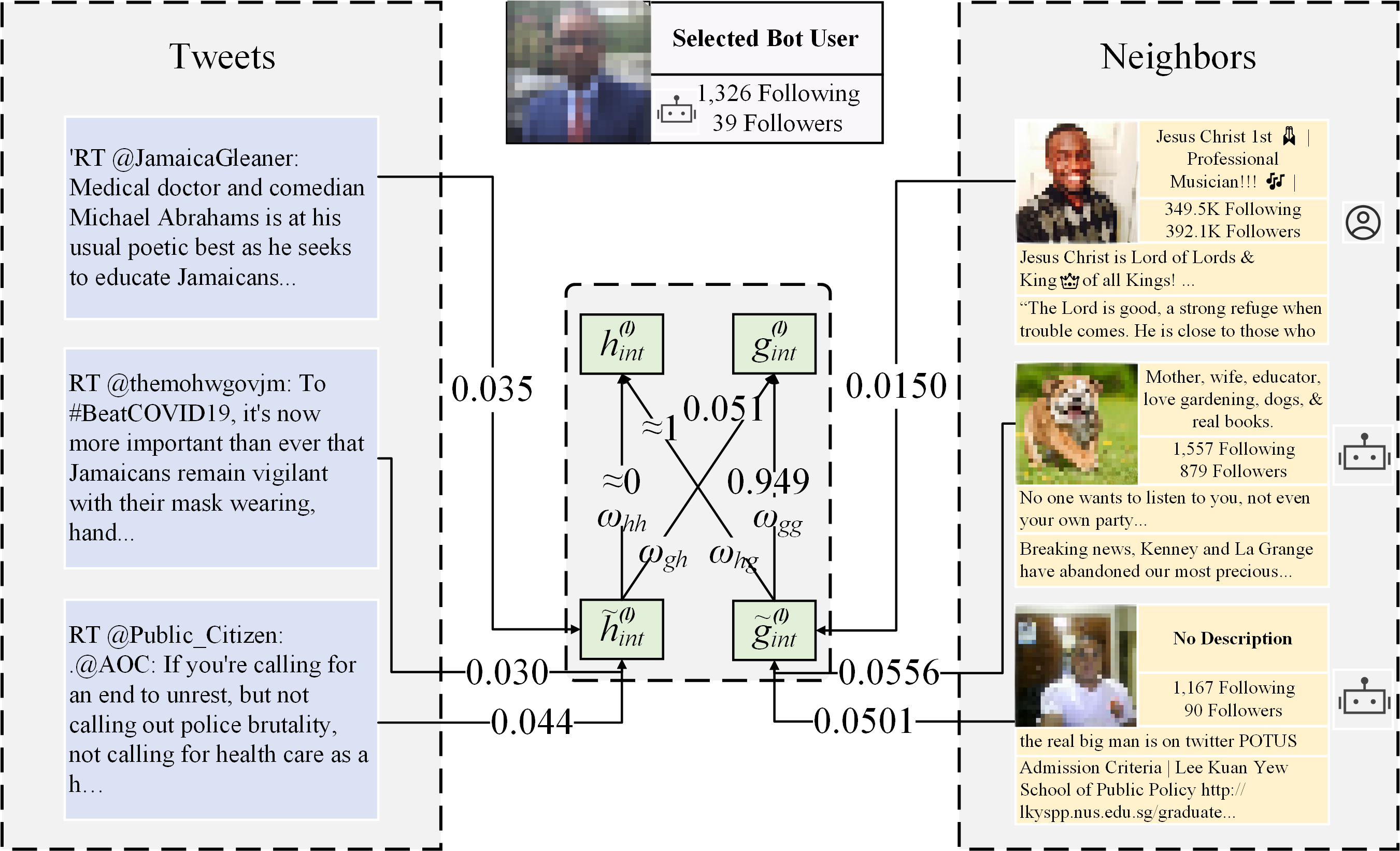}
    \caption{A sample user with its similarity weights inside the box in the middle. On the left are tweets with attention weights from the transformer in the text module. On the right are its neighbors with attention weights from multi-head attention in the graph module.}
    \label{fig:case}
\end{figure}

We study a specific Twitter user to explain how BIC exchanges information across two modalities and learns the relative importance to identify bots. For this user, we study its tweets and neighbors with the top-3 highest attention weight. We then derive similarity weights in Equation~(\ref{eq:s_weight}) to quantitatively analyze it. This user is visualized in Figure~\ref{fig:case}. We discovered that neighborhood information is more important in this cluster, due to more differences in attention weights of the selected bot's bot neighbors and human neighbors than attention weights of tweets. The conclusion is also reflected in similarity weights. The similarity weights of the original interaction representation from text modality are $0$ and $0.051$, while the similarity weights of the original interaction representation from graph modality are $1$ and $0.949$. The results further display the effectiveness of similarity-based interaction in that it indeed learns the emphasis on modalities.

\section{Related Work}
\subsection{Twitter-bot Detection}
\paragraph{Text-based Methods} Text-based methods adopt techniques in natural language processing to identify bots.~\citet{wei2019twitter} adopted multiple layers of bidirectional LSTM to conduct bot detection.~\citet{DBLP:conf/ijcai/StantonI19} proposed to leverage generative adversarial networks to detect spam bots.~\citet{hayawi2022deeprobot} adopted a variety of features and leveraged LSTM and dense layer to learn representations. Existing models can not capture the semantic consistency of users, which leads to failures to detect the advanced bots.
\paragraph{Graph-based Methods} Social network consist of rich information like social familiarity \citep{dey2017assessing, dey2018assessing}, attribution similarity \citep{peng2018anomalous}, and user interaction \cite{10.1145/1592665.1592675}. The graph constructs on Twittersphere help to detect bots.~\citet{feng2021satar} leveraged user neighbor information combined with tweet and profile information. Graph neural networks are utilized to improve the Twitter bot detectors and can achieve great performance~\citep{magelinski2020graph,DBLP:conf/data/DehghanSSDBMXKP22,yang2022rosgas}.~\citet{ali2019detect} used graph convolutional graph networks to aggregate user information.~\citet{feng2021botrgcn} constructed a heterogeneous graph and adopted relational graph convolutional graph networks to identify bots. Previous models leverage the text or graph modality alone without information interaction. We believe that exchanging modality information across two modalities can help improve performance. 

\subsection{Text-Graph Interaction}
Text information is the basis of natural language processing and pre-trained language models are the dominant framework in capturing text features~\citep{devlin-etal-2019-bert, DBLP:journals/corr/abs-1907-11692, lewis2020bart}. Meanwhile, graph neural networks are introduced to tackle NLP tasks, like fake news detection \citep{mehta-etal-2022-tackling}, dialogue state tracking \citep{feng-etal-2022-dynamic}, and machine translation \citep{xu-etal-2021-document-graph}.  As both pre-trained LMs and graph structure are proved to be effective, text-graph interaction was also widely used in the area of natural language processing. Some works interacted with two modalities hierarchically such as using encoded representations from knowledge graph to augment the textual representation~\citep{mihaylov2018knowledgeable, lin2019kagnet,yang2019enhancing}, or utilizing text representations to enhance the inferential capability of graph~\citep{feng2020scalable,lv2020graph}. More Recently, GreaseLM~\citep{DBLP:journals/corr/abs-2201-08860} proposed a model to allow two modalities to interact between layers by interaction nodes, in which truly deep interaction was achieved.


\section{Conclusion}
Twitter bot detection is a challenging task with increasing importance. To conduct a more comprehensive bot detection, we proposed a bot-detection model named BIC based. BIC interacts and exchanges information across text modality and graph modality by a text-graph interaction module. BIC contains a semantic consistency module that derives the inconsistency from tweets by the attention weight to identify advanced bots. We conducted extensive experiments on two widely used benchmarks to demonstrate the effectiveness of BIC in comparison to competitive baselines. Further experiments also bear out the effectiveness of modality interaction and semantic consistency detection. In the future, we plan to explore better interaction approaches.

\bibliography{anthology,custom}
\bibliographystyle{acl_natbib}

\appendix

\section{Limitations}
The BIC framework has two minor limitations:
\begin{itemize}[leftmargin=*]
    \item Our proposed BIC model utilizes representation from three different modalities, namely text, graph, and semantic consistency, and we introduce an interaction mechanism to allow information exchange between text and graph. However, whether interaction and information exchange is necessary among all three modalities is still an open question. We leave it to future work to study the necessary by introducing interaction modules.
    \item The new dataset we construct is limited to the topic of the pandemic while other popular topics are not considered. However, Twitter bots are likely to behave differently with different topics. We leave it to future works to analyze how current approaches perform against bots with different topics.
\end{itemize}

\section{Social Impact}
Our proposed BIC is a Twitter bot detection model that leverages text-graph interaction and semantic consistency modules. However, there are potential biases or discrimination that exist among the text, graph, or semantic consistency-based representation. For instance, some individuation users may be divided into the bot class for they may behave relevantly 'abnormal'. In conclusion, we suggest that the application of the Twitter bot detection model should be supervised by users and experts.

\section{Implementation Details}
\label{appendix:A}
We implement our framework with pytorch~\citep{paszke2019pytorch}, PyTorch geometric~\citep{fey2019fast}, and the transformer library from huggingface~\citep{wolf2019huggingface}. We limit each user's tweet number to 200, and for those who have posted fewer tweets, we bring their initial embeddings up to full strength with vectors made up of all zeros. 


\subsection{Hyperparamter Setting}
Table~\ref{tab:hyperparameter} presents the hyperparameter settings of BIC. For early stopping, we utilize the package provided by Bjarten\footnote{https://github.com/Bjarten/early-stopping-pytorch}.
\begin{table}
    \centering
    \caption{Hyperparameter settings of BIC.}
    \begin{tabular}{l c}
         \toprule[1.5pt] \textbf{Hyperparameter} & \textbf{Value} \\ \midrule[0.75pt]
         model layer count~$M$& 2 \\
         graph module input size & 768 \\
         graph module hidden size & 768 \\
         text module input size & 768\\
         text module hidden size & 768\\
         epoch & 30 \\
         early stop epoch & 10 \\
         batch size & 64 \\
         dropout & 0.5 \\
         learning rate & 1e-4 \\
         L2 regularization & 1e-5 \\
         lr\_scheduler\_patience & 5\\
         lr\_scheduler\_step & 0.1 \\
         Optimizer & RAdamW \\
         \bottomrule[1.5pt]
    \end{tabular}
    
    \label{tab:hyperparameter}
\end{table}

\subsection{Computation}
Our proposed method totally has 4.2M learnable parameters and 0.92 FLOPs\footnote{https://github.com/Lyken17/pytorch-OpCounter} with hyperparameters presented in Table~\ref{tab:hyperparameter}. Our implementation is trained on an NVIDIA GeForce RTX 3090 GPU with 24GB memory, which takes approximately 0.06 GPU hours for training an epoch.

\section{Baseline Details}
\begin{itemize}[leftmargin=*]
    \item\textbf{SATAR}~\citep{feng2021satar} leverages the tweet, profile, and neighbor information and employs a co-influence module to combine them. It pre-trains the model with the follower count and fine-tunes it to detect bots.
    \item\textbf{Botometer}~\citep{davis2016botornot} is a publicly available service that leverages thousands of features to evaluate how likely a Twitter account exhibits similarity to the known characteristics of typical bots.
    \item\textbf{Kudugunta \textit{et al.}}~\citep{kudugunta2018deep} subdivide bot-detection task to account-level classification and tweet-level classification. In the former, they combine synthetic minority oversampling (SMOTE) with undersampling techniques, and in the latter they propose an architecture that leverages a user's tweets.
    \item\textbf{Wei \textit{et al.}}~\citep{wei2019twitter} propose a bot detection model with a three-layer BiLSTM to encode tweets, before which pre-trained GloVe word vectors are used as word embeddings.
    \item\textbf{Alhosseini \textit{et al.}}~\citep{ali2019detect} utilize GCN to learn user representations from metadata such as user age, statuses\_count, account length name, followers\_count to classify bots.    \item\textbf{BotRGCN}~\citep{feng2021botrgcn} constructs a framework based on relational graph convolutional network~(R-GCN) by leveraging representatives derived from the combination of user tweets, descriptions, numerical and categorical property information.
    \item\textbf{Yang \textit{et al.}}~\citep{yang2020scalable} adopt random forest with account metadata for bot detection,  which is proposed to address the scalability and generalization challenge in Twitter bot detection.
    \item\textbf{RGT}~\citep{feng2022heterogeneity} leverages relation and influence heterogeneous graph network to conduct bot detection. RGT first learns users’ representation under each relation with graph transformers and then integrates representations with the semantic attention network.
\end{itemize}

\section{Evaluation Details}
\label{appendix:B}
We elaborate on the evaluation of our baselines here. For methods without text and graph modalities. \citet{lee2011long} adopt random forest classifier with Twitter bot features. \citet{yang2020scalable} adopt random forest with minimal account metadata. \citet{miller2014twitter} extract 107 features from a user’s tweet and metadata. \citet{cresci2016dna} encodes the sequence of a user's online activity with strings. Botometer~\citep{davis2016botornot} leverages more than one thousand features. All of them extract Twitter bot features, without dealing with these features in graph modality or text modality.

For methods with only text modality, SATAR~\citep{feng2021satar} leverages LSTM for its tweet-semantic sub-network. \citet{kudugunta2018deep} adopt deep neural networks for tackling user tweets. \citet{wei2019twitter} propose a model with a three-layer BiLSTM. All of them deal with user information in text modalities.

For methods with only graph modality, BotRGCN \citep{feng2021botrgcn} utilizes a relational graph convolutional network in its proposed framework. \citet{ali2019detect} adopt graph convolution network to learn user representations and classify bots. RGT~\citep{feng2022heterogeneity} leverages heterogeneous graph network to conduct bot detection. All of them deal with user information in graph modalities.

\section{Modality interaction additional study}
We conduct a qualitative experiment where we find that at least 54.8\% of bots~(51 / 93) that evaded the detection of only-text or only-graph models were captured by our proposed BIC, which also demonstrates BIC's effectiveness.

\section{Semantic consistency Study}
\begin{figure}
    \centering
    \includegraphics[width=0.9\linewidth]{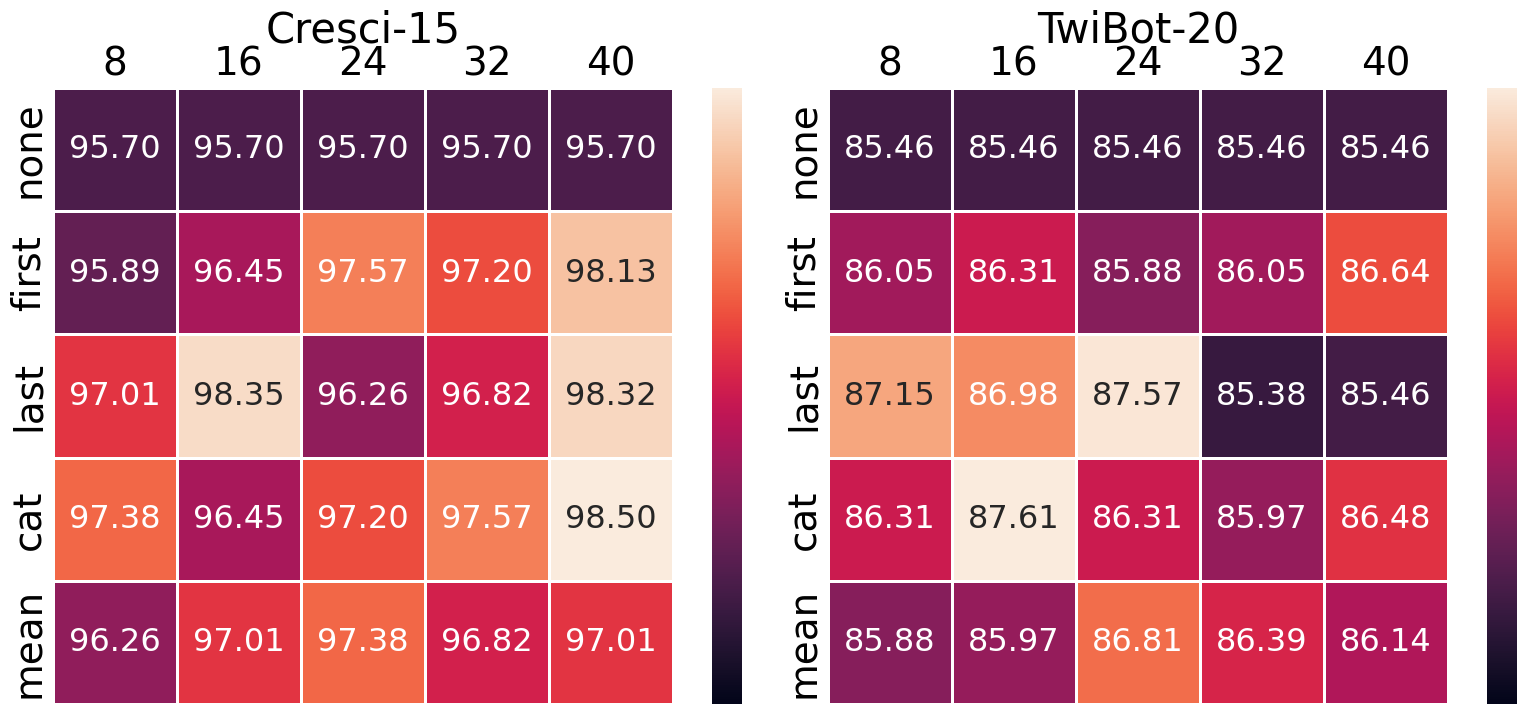}
    \caption{Accuracy of BIC with different settings of considering semantic consistency. The results illustrate the semantic consistency module can improve the performance.}
    \label{fig:sc_mode}
\end{figure}

\paragraph{Performance study}
To find how much semantic consistency detection helps the overall BIC performance with different parameter settings, we experiment with different semantic consistency layers, consistency matrix pooling sizes, and consistency vector aggregation manners. The results shown in Fig~\ref{fig:sc_mode} demonstrate that semantic consistency truly enhances the model performance. Although slight, differences are manifested in different parameter settings, which could be further studied.

\section{multi-task learning approach}
Apart from using interaction module to incorporate and exchange information between two modalities, we also consider multi-task learning approach which might be more straightforward and intuitive, and have similar effect. We conduct multi-task learning with both soft and hard parameters. The results shown in Table~\ref{tab:task} demonstrate that compared with multi-task learning, interaction module could better exchange information between two modalities and lead to a better performance.

\begin{table}
    \centering
     \caption{Performance of different task settings, where \textit{Multi-task~(hard)} or \textit{Multi-task~(soft)} refers to training regarding graph and text modalities as two different tasks with hard or soft parameter sharing. The results demonstrate that our proposed BIC and the modality interaction layer is empirically better at capturing the correlation between texts and networks for social media users. }
    \resizebox{\linewidth}{!}{
    \begin{tabular}{lcccc}
         \toprule[1.5pt]
         \multirow{2}{*}{\textbf{Methods}} & \multicolumn{2}{c}{\textbf{Cresci-15}} & \multicolumn{2}{c}{\textbf{TwiBot-20}}\\
         \cmidrule[0.75pt](lr){2-3}
         \cmidrule[0.75pt](lr){4-5}
         & \textbf{Accuracy} & \textbf{F1-score} & \textbf{Accuracy} & \textbf{F1-score}\\
         \midrule[0.75pt]
         Multi-task~(hard) &  $96.45$ & $97.72$&$84.62$& $86.54$ \\
         Multi-task~(soft) & $97.94$ & $98.39$& $84.45$& $85.82$ \\
         BIC &$\textbf{98.35}$ &$\textbf{98.71}$ & $\textbf{87.61}$& $\textbf{89.13}$ \\
         \bottomrule[1.5pt]
    \end{tabular}
    }
    \label{tab:task}
\end{table}

\section{Scientific Artifact}
The BIC model is implemented with the help of many widely-adopted scientific artifacts, including PyTorch~\citep{paszke2019pytorch}, NumPy~\citep{harris2020arraynumpy}, transformers~\citep{wolf2019huggingface}, sklearn~\citep{scikit-learn}, PyTorch Geometric~\citep{fey2019fast}. We commit to making our code and data publicly available to facilitate reproduction and further research.

\end{document}